\documentclass[letterpaper]{article} 
\usepackage{latex_sty/iclr2017_conference,times}
\usepackage[pass]{geometry}
\usepackage{url}
\usepackage[utf8]{inputenc}
\usepackage{xargs}                      
\usepackage{float}
\usepackage{tikz}
\usetikzlibrary{shapes,calc,decorations,plotmarks}

\usepackage[pdftex,
            pdfauthor={Leon~Sixt,~Benjamin~Wild,~\&~Tim~Landgraf},
            pdftitle={RenderGAN:~Generating~Realistic~Labeled~Data}
            ]{hyperref}

\usepackage{array}
\usepackage{calc}
\usepackage{multicol}
\usepackage{pifont}
\usepackage{amsfonts, amsopn, amsmath, amsthm, amssymb} 
\usepackage{comment}
\usepackage{lipsum}
\usepackage{etoolbox}
\usepackage{adjustbox}
\usepackage{bbm}
\usepackage[linewidth=1pt]{mdframed}
\usepackage{import}
\usepackage[toc,page]{appendix}
\usepackage{subcaption}

\graphicspath{{./images}{./images/plots}{./images/human_mhd_test}}

\IfFileExists{config_header.tex}{\newcommand{\decoderRenderGANMHD}{0.424}\newcommand{\decoderRealMHD}{0.956}\newcommand{\decoderRealRenderGANMHD}{0.416}\newcommand{\decoderHMThreeDMHD}{0.820}\newcommand{\decoderHMLightenMHD}{0.491}\newcommand{\decoderHMBGMHD}{0.505}\newcommand{\stagesBlurFactor}{0.90}\newcommand{\CVMHD}{1.08}}{}

\newtoggle{preview}
\toggletrue{preview}
\togglefalse{preview}

\newtoggle{paper}
\toggletrue{paper}

\iftoggle{preview}{
    \newcommand{\ifthesis}[1]{
        \begin{mdframed}[leftmargin=5pt,rightmargin=5pt,linecolor=red!30]
                \emph{\small Thesis only}\\
                #1
        \end{mdframed}
    }

    \newcommand{\ifpaper}[1]{
        \begin{mdframed}[leftmargin=5pt,rightmargin=5pt,linecolor=blue!30]
            \emph{\small Paper only:}\\
                #1
        \end{mdframed}
    }
    
    \newcommand{\we}{[we or I]}
    \newcommand{\We}{[We or I]}
}{
    \iftoggle{paper}{
        \newcommand{\ifpaper}[1]{#1}
        \newcommand{\ifthesis}[1]{}
        
        \newcommand{\we}{we }
        \newcommand{\We}{We }

    }{
        \newcommand{\ifpaper}[1]{}
        \newcommand{\ifthesis}[1]{#1}
        
        \newcommand{\we}{I }
        \newcommand{\We}{I }

    }
}
\DeclareMathOperator{\sigmoid}{sigmoid}

\usepackage{xargs}                      
\usepackage[colorinlistoftodos,prependcaption,textsize=tiny]{todonotes}

\newcommandx{\unsure}[2][1=]{\todo[inline,linecolor=red,backgroundcolor=red!25,bordercolor=red,#1]{#2}}
\newcommandx{\change}[2][1=]{\todo[linecolor=blue,backgroundcolor=blue!25,bordercolor=blue,#1]{#2}}
\newcommandx{\info}[2][1=]{\todo[linecolor=OliveGreen,backgroundcolor=OliveGreen!25,bordercolor=OliveGreen,#1]{#2}}
\newcommandx{\improvement}[2][1=]{\todo[linecolor=Plum,backgroundcolor=Plum!25,bordercolor=Plum,#1]{#2}}

\newcommand{\cn}[1]{{\color{red} \textsuperscript{[citation needed]}}}

\newboolean{show}
\setboolean{show}{true}
\ifshow
  
\else
  \excludecomment{questionize}
\fi

\DeclareMathOperator\highpass{highpass}

\let\pgfimageWithoutPath\pgfimage
\renewcommand{\pgfimage}[2][]{\pgfimageWithoutPath[#1]{images/plots/#2}}

\newcommand{\fillabstract}{
Deep Convolutional Neuronal Networks (DCNNs) are showing remarkable performance
on many computer vision tasks.  Due to their large parameter space, they require
many labeled samples when trained in a supervised setting. The costs of
annotating data manually can render the use of DCNNs infeasible.  We present
a novel framework called RenderGAN that can generate large amounts of realistic,
labeled images by combining a 3D model and the Generative Adversarial Network
framework. In our approach, image augmentations (e.g. lighting, background, and
detail) are learned from unlabeled data such that the generated images are
strikingly realistic while preserving the labels known from the 3D model.  We
apply the RenderGAN framework to generate images of barcode-like markers that
are attached to honeybees. Training a DCNN on data generated by the
RenderGAN yields considerably better performance than training it on
various baselines.  }

\title{RenderGAN: Generating Realistic Labeled Data}
\author{Leon Sixt, Benjamin Wild, \& Tim Landgraf \\
Fachbereich Mathematik und Informatik\\
Freie Universität Berlin\\
Berlin, Germany \\
{\small \texttt{\{leon.sixt, benjamin.wild, tim.landgraf\}@fu-berlin.de} }\\
}

\begin{document}

\maketitle

\begin{abstract}
\fillabstract
\end{abstract}

%

\section{Introduction}

When an image is taken from a real world scene, many factors determine the final
appearance: background, lighting, object shape, position and orientation of the
object, the noise of the camera sensor, and more.  In computer vision,
high-level information such as class, shape, or pose is reconstructed  from raw
image data.  Most real-world applications require the reconstruction to be
invariant to noise, background, and lighting changes.

In recent years, deep convolutional neural networks (DCNNs) advanced to the
state of the art in many computer vision tasks \citep{Krizhevsky2012, He2015,
Razavian2014}. More training data usually increases the performance of DCNNs.
While image data is mostly abundant, labels for supervised training must often
be created manually -- a time-consuming and tedious activity. For complex
annotations such as human joint angles, camera viewpoint or image segmentation,
the costs of labeling can be prohibitive.

In this paper, we propose a method to drastically reduce the costs of labeling
such that we can train a model to predict even complex sets of labels. We
present a generative model that can sample from the joint distribution of labels
and data. The training procedure of our model does not require any manual
labeling. We show that the generated data is of high quality and can be used to
train a model in a supervised setting, i.e.\ a model that maps from real samples
to labels, without using any manually labeled samples.

We propose two modifications to the recently introduced GAN framework
\citep{Goodfellow2014}. First, a simple 3D model is embedded into the generator
network to produce samples from corresponding input labels. Second, the
generator learns to add missing image characteristics to the model output using
a number of parameterized augmentation functions. In the adversarial training we
leverage large amounts of unlabeled image data to learn the particular form of
blur, lighting, background and image detail. By constraining the augmentation
functions we ensure that the resulting image still represents the given set of
labels. The resulting images are hard to distinguish from real samples and can
be used to train a DCNN to predict the labels from real input data.

\begin{figure*}[t]
    \centering
    \begin{subfigure}[b]{.35\textwidth}
        \centering
        \def\svgwidth{0.60\textwidth}
        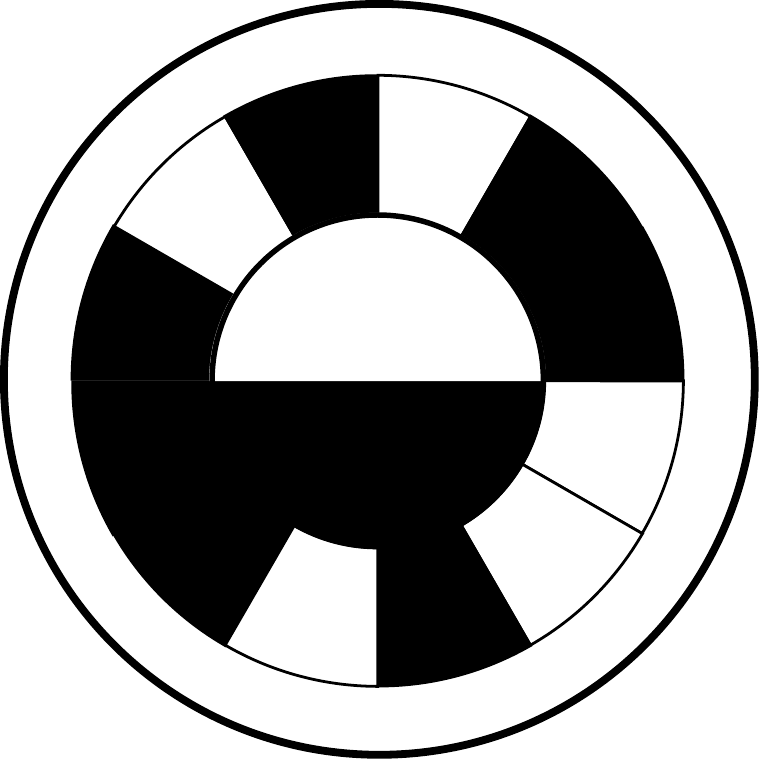
        \caption{Tag structure}
        \label{fig:subplot_fake}
    \end{subfigure}
    \begin{subfigure}[b]{.50\textwidth}
        \centering
        \includegraphics[width=0.55\textwidth]{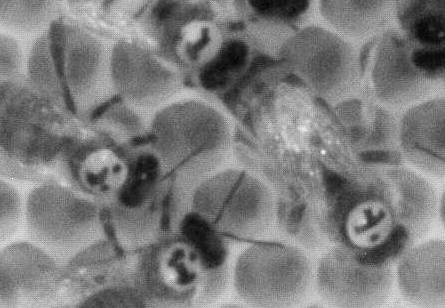}
        \caption{Tagged bees in the hive}
        \label{fig:subplot_real}
    \end{subfigure}
    \caption{
        \textbf{(a)} The tag represents a unique binary code (cell 0 to 11) and encodes
        the orientation with the semicircles 12 and 13.
        The red arrow points towards the head of the bee.
        This tag encodes the id \texttt{100110100010}.
        \textbf{(b)} Cutout from a high-resolution image. 
    }
    \label{fig:tag_beesbook}
\end{figure*}

\begin{figure}[b]
    \centering
    \includegraphics[width=0.8\textwidth]{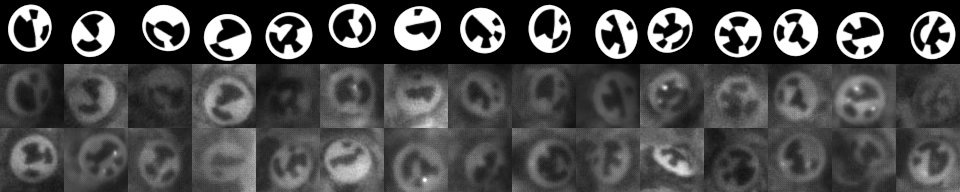}
    \caption{
        \textbf{First row:} Images from the 3D model without augmentation.
        \textbf{Below:} Corresponding images from the RenderGAN.
        \textbf{Last row:} Real images of bee's tags.
    }
    \label{fig:tag3dfake}
\end{figure}

The RenderGAN framework was developed to solve the scarcity of labeled data in
the BeesBook project \citep{Wario2015} in which we analyze the social behavior
of honeybees.  A barcode-like marker is attached to the honeybees' backs for
identification (see Fig.~\ref{fig:tag_beesbook}).  Annotating this data is
tedious, and therefore only a limited amount of labeled data exists. A 3D model
(see the upper row of Fig.~\ref{fig:tag3dfake}) generates a simple image of the
tag based on position, orientation, and bit configuration. The RenderGAN then
learns from unlabeled data to add lighting, background, and image details.

Training a DCNN on data generated by the RenderGAN yields considerably better
performance compared to various baselines. We furthermore include a previously
used computer vision pipeline in the evaluation. The networks' detections are
used as feature to track the honeybees over time.  When we use detections from
the DCNN instead of the computer vision pipeline, the accuracy of assigning the
true id increases from 55\% to 96\%.

Our contributions are as follows. We present an extension of the
GAN framework that allows to sample from the joint distribution of data and
labels. The generated samples are nearly indistinguishable from real data for
a human observer and can be used to train a DCNN end-to-end to classify real
samples. In a real-world use case, our approach significantly outperforms
several baselines. Our approach requires no manual labeling. The
simple 3D model is the only form of supervision.

\section{Related work}

There exists multiple approaches to reduce the costs associated with labeling.

A common approach to deal with limited amount of labels is data augmentation
\citep[Chapter 7.4]{Goodfellow2016}.  Translation, noise, and other deformations
can often be applied without changing the labels, thereby effectively increasing
the number of training samples and reducing overfitting.

DCNNs learn a hierarchy of features -- many of which are applicable to related
domains \citep{Yosinski2014}. Therefore, a common technique is to pre-train a
model on a larger dataset such as ImageNet \citep{Deng2009} and then fine-tune
its parameters to the task at hand \citep{Malik2014,Long2015,Razavian2014}.
This technique only works in cases where a large enough related dataset exists.
Furthermore, labeling enough data for fine-tuning might still be costly.

If a basic model of the data exists (for example, a 3D model of the human body),
it can be used to generate labeled data.  \cite{Peng2015} generated training
data for a DCNN with 3D-CAD models.  \cite{Su2015} used 3D-CAD models from large
online repositories to generate large amounts of training images for the
viewpoint estimation task on the PASCAL 3D+ dataset \citep{Xiang2014}.
\cite{Massa2015} are matching natural images to 3D-CAD models with features
extracted from a DCNN.  \cite{Richter2016ECCV} and \cite{RosCVPR16} used 3D game
engines to collect labeled data for image segmentation.  However, the explicit
modeling of the image acquisition physics (scene lighting, reflections, lense
distortions, sensor noise, etc.) is cumbersome and might still not be able to
fully reproduce the particularities of the imaging process such as unstructured
background or object specific noise.  Training a DCNN on generated data that
misses certain features will result in overfitting and poor performance on the
real data.

\begin{figure}[t]
    \centering
        \begin{tikzpicture}
    [
     thick,
     input/.style={rectangle,fill=none},
     network/.style={rectangle,inner sep=6pt,fill=blue!10,draw=black,font=\large,align=center},
    ]
    \coordinate (zc) at (0, 0);
    \coordinate (dc) at ($ (zc) + (4.2, 0.7)$);
    \coordinate (rc) at ($ (zc) + (0, 1.4) $);
    \coordinate (gc) at ($ (zc) + (1.5, 0) $);
    \coordinate (fc) at ($ (gc) + (1.5, 0) $);
    \node[input] (r) at (rc)  {real data};
    \node[input] (z) at (zc) {noise};
    \node[input] (o) at ($ (dc) + (1.5, 0) $)  {$0\text{-}1$};
    \node[network] (g) at (gc)  {$G$};
    \node[network] (d) at (dc) {$D$};
    \draw[]
    (d) -- (o)
    (r.east) -- ($(fc) + (0, 1.4) $);
    \draw[]
    (g.east) --  (fc) node[pos=0.6,fill=white]{fake};

    \draw[->]
    (z) -- (g);
    \draw[->]
    ($(fc) + (0, 1.4) $) to[out=0, in=180] ($ (d.west) + (0, 0.2)$);
    \draw[->]
    (fc) to[out=0, in=180]  ($ (d.west) + (0, -0.2)$);

\end{tikzpicture}
    \caption[foo]{
        Topology of a GAN. The discriminator network $D$ is trained
        to distinguish between "fake" and real data.
        The generator network $G$ receives a random vector as input.
        $G$ is optimized to maximize the chance of the discriminator making a mistake.
    }
    \label{fig:gan}
\end{figure}
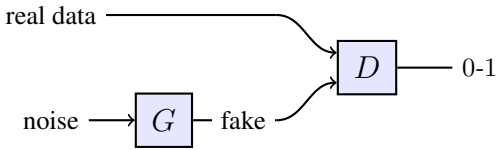

Generative Adversarial Networks (GAN) (see Fig.~\ref{fig:gan}) can learn to
generate high-quality samples \citep{Goodfellow2014}, i.e.\ sample from the data
distribution $p(x)$.  \cite{Denton2015} synthesized images with a GAN on the
CIFAR dataset \citep{Krizhevsky2009}, which were hard for humans to distinguish
from real images.  While a GAN implicitly learns a meaningful latent embedding
of the data \citep{Radford2015}, there is no simple relationship between the
latent dimensions and the labels of interest. Therefore, high-level information
can’t be inferred from generated samples. cGANs are an extension of GANs to
sample from a conditional distribution given some labels, i.e.\ $p(x|l)$.
However, training cGANs requires a labeled dataset.  \cite{Springenberg2015}
showed that GANs can be used in a semi-supervised setting but restricted their
analysis to categorical labels.  \cite{Wang2016} trained two separate GANs, one
to model the object normals and another one for the texture conditioned on the
normals.  As they rely on conditional GANs, they need large amounts of labeled
data.  \cite{Chen2016} used an information theoretic to disentangle the
representation.  They decomposed the representation into a structured and
unstructured part.  And successfully related on a qualitative level the
structured part to high-level concepts such as camera viewpoint or hair style.
However, explicitly controlling the relationship between the latent space and
generated samples \emph{without using labeled data} is an open problem, i.e.\
sampling from $p(x,l)$ without requiring labels for training.

\section{RenderGAN}
Most supervised learning tasks can be modeled as a regression problem, i.e.\
approximating a function $\hat f: \mathbb{R}^n \mapsto L$ that maps from data space $\mathbb{R}$ to
label space $L$.  We consider $\hat f$ to be the best available
function on this particular task. Analogous to ground truth data, one could call
$\hat f$ the ground truth function.

In the RenderGAN framework, we aim to solve the inverse problem to this
regression task: generate data given the labels.  This is achieved by embedding
a simple 3D model into the generator of a GAN.  The samples generated by the
simple model must correspond to the given labels but may lack many factors of
the real data such as background or lightning.  Through a cascade of
augmentation functions, the generator can adapt the images from the 3D model to
match the real data.

We formalize image augmentation as a function $\phi(x, d)$, which modifies the
image $x$ based on the augmentation parameter $d$ (a tensor of any rank). The
augmentation must preserve the labels of the image $x$.  Therefore, it must hold
for all images $x$ and all augmentations parameters $d$:

\begin{equation}
    \label{eq:augmentation}
    \hat f \left( \phi(x, d) \right) = \hat f(x)
\end{equation}

The augmentation function must furthermore be differentiable w.r.t. $x$ and $d$ as
the gradient will be back-propagated through $\phi$ to the generator.
Image augmentations such as lighting, surrounding, and noise do preserve the
labels and fit this definition. \We will provide appropriate definitions of $\phi$ for
the mentioned augmentations in the following section.

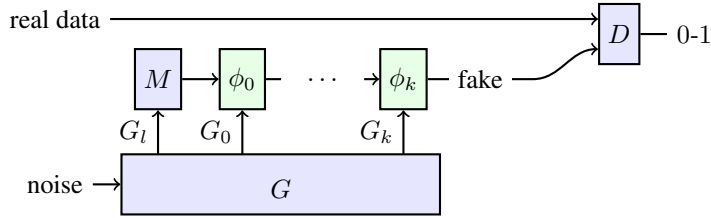
\begin{figure}[t]
    \centering
    \def\bbDistAug{1.4}
\def\bbArrow{1.0}
\begin{tikzpicture}
    [
     thick,
     input/.style={rectangle,fill=none},
     network/.style={rectangle,minimum height=0.8cm, minimum width=0.5cm,fill=blue!10,draw=black,align=center},
     combination/.style={network,rectangle,fill=green!10,draw=black,align=center},
    ]
    \coordinate (zc) at (0, 0);
    \coordinate (gc) at ($ (zc) + (3, -0.5)$);
    \coordinate (dc) at ($ (zc) + (7.5, 1.5)$);
    \coordinate (rc) at ($ (zc) + (0, 1.7) $);
    \node[input] (r) at (rc)  {real data};
    \node[input] (z) at ($ (zc) + (0, -0.5)$) {noise};
    \node[input] (out) at ($ (dc) + (1, 0) $)  {$0\text{-}1$};
    \node[network,align=center,text width=4.0cm, text height=1em] (g) at (gc)  {$G$};
    \node[network] (d) at (dc) {$D$};
    \node[network] (m) at ($ (g.west) + (0.5, \bbDistAug)$) {$M$};
    \node[combination] (p0) at ($ (m.east) + (0.8, 0)$) {$\phi_0$};
    \node[combination] (pk) at ($ (g.east) + (- 0.5, \bbDistAug)$) {$\phi_k$};
    \node[network, fill=none, draw=none] (p1) at ($ 0.5*(p0.east) + 0.5*(pk.west)$) {$\ldots$};
    \draw[]
    (d) -- (out)
    (p0.east) -- ($(p0.east) + (0.2, 0)$)
    (pk.east) --  ($(pk.east) + (1.4,0)$) node[midway,fill=white]{fake}
    ;
    \draw[->]
    (z) edge (g);
    \draw[->] ($(m) - (0, \bbArrow)$) -- (m) node[midway,left] {$G_l$};
    \draw[->] ($(p0) - (0, \bbArrow)$) -- (p0) node[midway,left] {$G_0$};
    \draw[->] ($(pk) - (0, \bbArrow)$) -- (pk) node[midway,left] {$G_k$};
    \draw[->]
    (m.east) edge (p0.west)
    ($(pk.west) - (0.2, 0)$) edge (pk.west)
    (r.east) edge ($ (d.west) + (0, 0.2) $)
    ($(pk.east) + (1.4,0)$) to[out=0, in=180] ($(d.west) - (0, 0.2) $)
    ;

\end{tikzpicture}
    \caption[foo]{
        The generator $G$ cannot directly produce samples.
        Instead, $G$ has to predict parameters $G_l$ for the 3D model $M$.
        The image generated by $M$ is then modified through the augmentation functions $\phi_i$
        parameterized by $G_i$ to match the real data.
    }
    \label{fig:rendergan}
\end{figure}

If appropriate augmentation functions are found that can model the missing
factors and are differentiable, we can use the GAN framework to find
parameters that result in realistic output images.
Multiple augmentation functions can be combined to perform a more complex
augmentation. Here, we will consider multiple augmentation functions applied
sequentially, i.e.\ we have $k$ augmentation functions $\phi_i$ and $k$
corresponding outputs $G_i$ from the generator.  The output of the previous
augmentation function is the input to the next one. Thus, we can write the
generator given some labels l as:
\begin{equation}
    g(z, l) = \phi_k(\phi_{k-1}(\ldots \phi_0(M(l), G_0(z)) \ldots, G_{k-1}(z)), G_k(z))
\end{equation}
where M is the 3D model.
We can furthermore learn the label distribution with the generator. As the discriminator loss must be backpropagated through the 3D model M, it must be differentiable. This can be achieved by emulating the 3D model with a neural network \citep{Dosovitskiy2014}.
The resulting generator g(z) can be written as (see Fig.~\ref{fig:rendergan} for a visual interpretation):
\begin{equation}
    \label{eq:rendergan}
    g(z) = \phi_k(\phi_{k-1}(\ldots \phi_0(M(G_l(z)), G_0(z)) \ldots, G_{k-1}(z)), G_k(z))
\end{equation}
As any differentiable function approximator can be employed in the GAN framework,
the theoretical properties still hold.  The training is carried out as in the
conventional GAN framework. In a real application, the augmentation functions might
restrict the generator from converging to the data distribution.

\label{sec:rendergan}

If the training converges, we can collect generated realistic data with $g(z)$
and the high-level information captured in the 3D model with $G_l(z)$. We can
now train a supervised learning algorithm on the labeled generated data
$\left(G_l\left(z\right), g\left(z\right)\right)$ and solve the regression task
of approximating $\hat f$ without depending on manual labels.

\section{Application to the BeesBook project}
\label{sec:applicationRendergan}

\begin{figure}[h]
    \centering
    \input{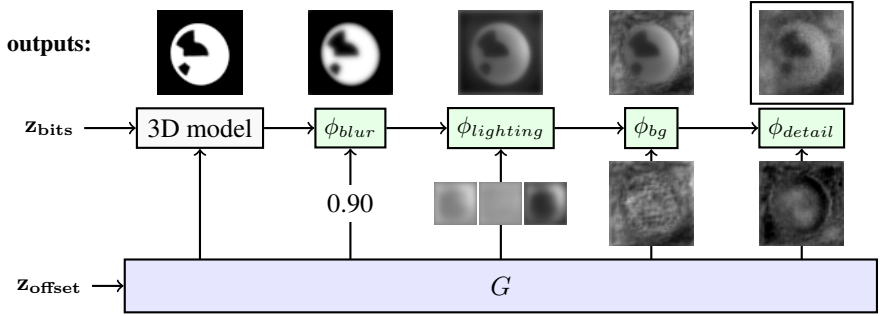}
    \caption{Augmentation functions of the RenderGAN applied to the BeesBook project.
        The arrows from $G$ to the augmentation functions $\phi$ depict
        the inputs to the augmentation functions. The generator provides the position and
        orientations to the 3D model, whereas the bits are sampled uniformly.
        On top, the output of each stage is shown. The output of $\phi_{detail}$ is forwarded to the discriminator.}
    \label{fig:bb_rendergan}
\end{figure}

In the BeesBook project, we aim to understand the complex social behavior
of honey bees.  For identification, a tag with a binary code is attached to the back of the bees.

The orientations in space, position, and bits of the tags are required to track
the bees over time. Decoding this information is not trivial: the bees are
oriented in any direction in space. The tag might be partially occluded.
Moreover, spotlights on the tag can sometimes even confuse humans. A previously
used computer vision pipeline did not perform reliably. Although we invested a
substantial amount of time on labeling, a DCNN trained on this data did not
perform significantly better (see Sec.~\ref{sec:rendergan}). We therefore
wanted to synthesize labeled images which are realistic enough to train an
improved decoder network.

Following the idea outlined in section~\ref{sec:rendergan}, we created a simple
3D model of a bee marker.  The 3D model comprises a mesh which represents the
structure of the marker and a simple camera model to project the mesh to the
image plane. The model is parameterized by its position, its pitch, yaw and
roll, and its ID.  Given a parameter set, we obtain a marker image, a background
segmentation mask and a depth map. The generated images lack many important
factors:  blur, lighting, background, and image detail (see
Fig.~\ref{fig:tag3dfake}). A DCNN trained on this data does not generalize well
(see Sec.~\ref{sec:results}).

Over the last years we collected a large amount of unlabeled image data. We
successfully augmented the 3D model using this dataset, as described below.

We trained a neural network to emulate the 3D model. Its outputs are
indistinguishable from the images of the 3D model. The discriminator error can
now be backpropagated through the 3D model which allows the generator to also
learn the distributions of positions and orientations of the bee marker. The IDs
are sampled uniformly during training. The weights of the 3D model network are
fixed during the RenderGAN training.

\We apply different augmentation functions that account for blur, lighting,
background, and image detail. The output of the 3D model and of each
augmentation function is of shape $(64, 64)$ and in the range $[-1, 1]$. In
Fig.~\ref{fig:bb_rendergan}, the structure of the generator is shown.

\textbf{Blurriness:} The 3D model produces hard edges, but the images of the
real tags show a broad range of blur.  The generator produces a scalar
$\alpha \in [0, 1] $ per image that controls the blur.
\begin{align}
    \phi_{blur}(x, \alpha) = (1 - \alpha) \left(x - b_\sigma\left(x\right)\right)  + b_\sigma(x)
\end{align}
where  $b_\sigma(x) = x*k_\sigma$ denotes convolving the image $x$
with a Gaussian kernel $k_\sigma$ of scale $\sigma$.
The implementation of the blur function is inspired by Laplacian pyramids
\citep{Burt1983}.
As required for augmentation functions,
the labels are preserved, because we limit the maximum amount of blur by picking
$\sigma  = 2$. $\phi_{blur}$ is also differentiable w.r.t the inputs $\alpha$ and $x$.

\textbf{Lighting of the tag:} The images from the 3D model are binary. In real images, tags
exhibit different shades of gray. \We model the lighting by a smooth scaling and shifting of the pixel intensities.
The generator provides three outputs for the lighting: scaling of black parts $s_b$,
scaling of white parts $s_w$ and a shift $t$. All outputs have the same dimensions as the image $x$.
An important invariant is that the black bits of the tag must stay darker than the white bits.
Otherwise, a bit could flip, and the label would change.
By restricting the scaling $s_w$ and $s_b$ to be between 0.10 and 1, \we ensure that this invariant holds.
The lighting is locally corrolated and should cause smooth changes in the image.
Hence, Gaussian blur $b(x)$ is applied to $s_b$, $s_w$, and $t$.
\begin{equation}
    \phi_{lighting}(x, s_w, s_b, t) = x \cdot b(s_w) \cdot W(x) +
                                       x \cdot b(s_b) \cdot (1-W(x))  + b(t)
\end{equation}
The segmentation mask $W(x)$ is one for white parts and zero for the black part of the image.
As the intensity of the input is distributed around -1 and 1,
we can use thresholding to differentiate between black and white parts.

\textbf{Background}:
The background augmentation can change the background pixels arbitrarily.
A segmentation mask $B_x$ marks the background pixels of the image $x$ which are
replaced by the pixels from the generated image $d$.
\begin{equation}
    \phi_{bg}(x, d) = x\cdot \left(1-B_x\right) + d \cdot B_x
\end{equation}
The 3D model provides the segmentation mask. As $\phi_{bg}$ can only change background
pixels, the labels remain unchanged.

\textbf{Details:} In this stage, the generator can add small details to the
whole image including the tag.
The output of the generator $d$ is passed through a high-pass filter
to ensure that the added details are small enough not to flip a bit.
Furthermore, $d$ is restricted to be in $[-2, 2]$ to make sure the generator cannot
avoid the highpass filter by producing huge values. With the range $[-2, 2]$, the generator has the possibility to
change black pixels to white, which is needed to model spotlights.

\begin{equation}
    \phi_{detail}(x, d) = x + \highpass(d)
\end{equation}
The high-pass is implemented by taking the difference between the image
and a blurred version of the image ($\sigma = 3.5$).
As the spotlights on the tags are only a little smaller than the bits, we
increase its slope after the cutoff frequency by repeating the high-pass filter
three times.

The image augmentations are applied in the order as listed above: $\phi_{detail}
\circ  \phi_{background} \circ \phi_{lighting} \circ \phi_{blur} $.
Please note that there exist parameters to the augmentation functions
that could change the labels. As long as it is guaranteed that such augmentations
will result in unrealistic looking images, the generator network will learn to avoid them.
For example, even though the detail augmentation could be used
to add high-frequency noise to obscure the tag,
this artifact would be detected by the discriminator.

\textbf{Architecture of the generator:} The generator network has to produce
outputs for each augmentation function.  We will outline only the most important
parts. See our code available online for all the details of the networks\footnote{\url{https://github.com/berleon/deepdecoder}}.
The generator starts with a small network consisting of dense layers, which predicts the
parameters for the 3D model (position, orientations).
The output of another dense layer is reshaped and used as starting block for
a chain of convolution and upsampling layers.
\We found it advantageous to merge a depth map of the 3D model into the generator as especially
the lighting depends on the orientation of the tag in space.
The input to the blur augmentation is predicted by reducing an intermediate
convolutional feature map to a single scalar.
An additional network is branched off to predict the input to the lighting augmentation.
For the background generation, the output of the lighting network is merged back into the main generator network
together with the actual image from the 3D model.

For the discriminator architecture, we mostly rely on the architecture given by
\cite{Radford2015}, but doubled the number of convolutional layers and added
a final dense layer. This change improved the quality of the generated images.

\textbf{Clip layer:}
Some of the augmentation parameters have to be restricted to a range of values
to ensure that the labels remain valid.  The training did not converge when
using functions like $\tanh$ or $\sigmoid$ due to vanishing gradients.  We are
using a combination of clipping and activity regularization to keep the output
in a given interval $[a, b]$.  If the input $x$ is out of bounds, it is clipped
and a regularization loss $r$ depending on the distance between $x$ and the
appropriate bound is added.
\begin{align}
      r(x) &= \begin{cases}
        \gamma||x - a||_{1} \quad  & \text{if } x < a  \\
                0 \quad  & \text{if } a \le x \le b \\
                \gamma||x - b||_{1} \quad  &\text{if } x > b
            \end{cases} \\
      f(x) &= \min(\max(a, x), b)
\end{align}
With the scalar $\gamma$, the weight of the loss can be adapted. For us $\gamma
= 15$ worked well. If $\gamma$ is chosen too small, the regularization loss
might not be big enough to move the output of the previous layer towards the
interval $[a, b]$. During training, we observe that the loss decreases to a
small value but never vanishes.

\begin{figure}[t]
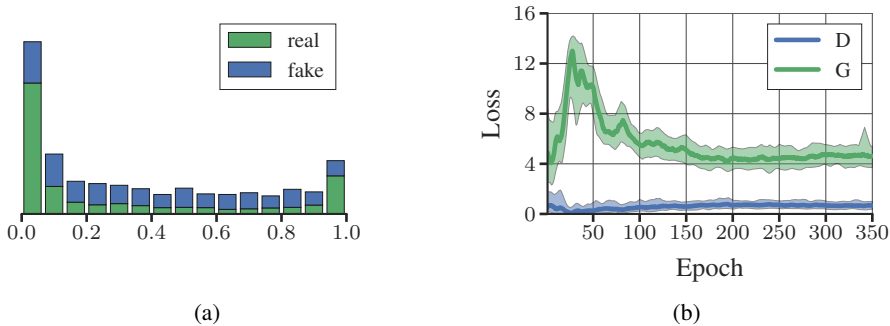

  \centering
    \begin{subfigure}[t]{.45\textwidth}
        \centering
        \input{images/plots/discriminator_score.pgf}
        \caption{}
        \label{fig:dscore}
    \end{subfigure}
    \begin{subfigure}[t]{.45\textwidth}
        \centering
        \input{images/plots/gan_history.pgf}
        \caption{}
        \label{fig:gan_history}
    \end{subfigure}
  \caption{
      \textbf{(a)} Histogram of the discriminator scores on fake and real samples.
      \textbf{(b)} Losses of the generator (G) and discriminator (D).
    }
\end{figure}

\textbf{Training:}
We train generator and discriminator as in the normal GAN setting.
We use 2.4M unlabeled images of tags to train the RenderGAN.
We use Adam \citep{Kingma2014} as an optimizer with a starting learning rate of
$0.0002$, which we reduce in epoch $200$, $250$, and $300$ by a factor of
$0.25$. In Fig.~\ref{fig:gan_history} the training loss of the GAN is shown.
The GAN does not converge to the point where the discriminator can no longer separate
generated from real samples. The augmentation functions might restrict the
generator too much such that it cannot model certain properties.
Nevertheless, it is hard for a human to distinguish the generated from real images.
In some cases, the generator creates unrealistic high-frequencies artifacts.
The discriminator unfailingly assigns a low score to theses images. We can therefore
discard them for the training of the supervised
algorithm. More generated images are shown in Appendix~\ref{appendix:generated_images}.
In Fig.~\ref{fig:rendergan_interpolation}, we show random points in the latent
space, while fixing the tag parameters. The generator indeed learned to model
the various lighting conditions, noise intensities, and backgrounds.

\begin{figure}[t]
    \centering
    \includegraphics[width=0.8\linewidth]{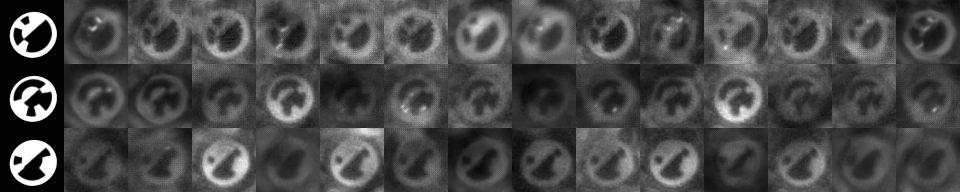}
    \caption{Random points in the z-space given the tag parameters}
    \label{fig:rendergan_interpolation}
\end{figure}

\section{Results}
\label{sec:results}

\begin{table}[b]
    \centering
    \caption{Datasets created with learned representations and hand-designed augmentations}
    \label{tab:handmade_data}
    \begin{tabular}{c | c | c }
        Name  & Learned  & Hand-Designed \\ \hline
        HM 3D &  3D model & blur, lighting, background, noise, spotlights \\  \hline
        HM LI &  3D model, blur, lighting & background, noise, spotlights \\ \hline
        HM BG &  3D model, blur, lighting, background & noise, spotlights \\
    \end{tabular}
\end{table}

We constructed the RenderGAN to generate labeled data. But does a DCNN trained
with the RenderGAN data perform better than one trained on the limited amounts
of real data? And are learned augmentations indeed needed or do simple
hand-designed augmentation achieve the same result?  The following paragraphs
describe the different datasets used in the evaluation.  We focus on the
performance of a DCNN on the generated data. Thus, we do not compare our method
to conventional GANs as those do not provide labels and are generally hard to
evaluate.

\textbf{Data from the \emph{RenderGAN}}:  We generate 5 million tags with the RenderGAN framework.
Due to the abundance, one training sample is only used twice during training. It is not further
augmented.

\textbf{\emph{Real} Data}: The labels of the real data are extracted from ground truth data that was
originally collected to evaluate bee trajectories.  This ground truth data
contains the path and id of each bee over multiple consecutive frames.  Data
from five different time spans was annotated -- in total 66K tags.
As the data is correlated in time (same ids, similar lighting conditions), we
assign the data from one time span completely to either the train or test set.
The data from three time spans forms the train set (40K). The test set (26K)
contains data from the remaining two time spans.  The ground truth data lacks
the orientation of the tags, which is therefore omitted for this evaluation.
Due to the smaller size of the real training set, we augment it with random
translation, rotation, shear transformation, histogram scaling, and noise
(see Appendix~\ref{appendix:augmentation_real} for exact parameters).

\textbf{\emph{RenderGAN + Real}}: We also train a DCNN on generated and
real data which is mixed at a 50:50 ratio.

\textbf{Handmade augmentations:}
We tried to emulate the augmentations learned by the RenderGAN by hand.
For example, we generate the background by an image pyramid where the
pixel intensities are drawn randomly. We model all effects, i.e.\ blur, lighting,
background, noise and spotlights (see Appendix~\ref{appendix:handmade} for
details on their implementation).
We apply the handmade augmentation to different learned representations  of the RenderGAN,
e.g.\ we use the learned lighting representation and add the remaining effects
such as background and noise with handmade augmentations (\emph{HM LI}). See
Table~\ref{tab:handmade_data} for the different combinations of learned representations and hand designed augmentations.

\textbf{Computer vision pipeline \emph{CV} :} The
previously used computer vision pipeline \citep{Wario2015} is based on manual
feature extraction. For example, a modified Hough transformation to find
ellipses. The MHD obtained by this model is only a rough estimate given that the computer vision
pipeline had to be evaluated and fine-tuned on the same data set due to label scarcity.

\textbf{Training setup}: An epoch consists of 1000 batches \'a 128 samples.  We
use early stopping to select the best parameters of the networks.  For the
training with generated data, we use the real training set as the validation
set. When training on real data, the test set is also used for validation.  We
could alternatively reduce the real training set further and form an extra
validation set, but this would harm the performance of the DCNN trained on the
real data. We use the 34-layer ResNet architecture \citep{He2015} but start with
16 feature maps instead of 64. The DCNNs are evaluated on the mean Hamming
distance (MHD) i.e.\ the expected value of bits decoded wrong. Human experts can
decode tags with a MHD of around 0.23.

\newcommand{\includestrip}[1]{
    \begin{minipage}{.68\textwidth}
        \includegraphics[width=\linewidth]{images/plots/training_samples_strip_#1.png}
    \end{minipage}
}

\begin{table}[t]
    \centering
    \caption{Comparison of the mean Hamming distance (MHD) on the different data sets.
    More samples of the training data can be found in Appendix~\ref{appendix:training_samples}.}
    \label{tab:mhd}
    \begin{tabular}{
             >{\centering\arraybackslash}m{5em}|
             >{\centering\arraybackslash}m{2em}|
             c
            }
        Data        & MHD                   & Training Data                 \\ \hline
        Real        & \decoderRealMHD       & \includestrip{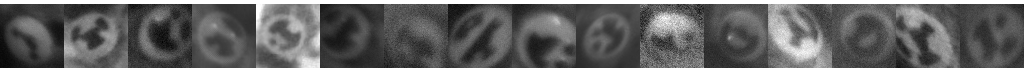}           \\ \hline
        HM 3D       & \decoderHMThreeDMHD   & \includestrip{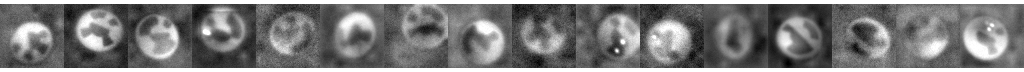}          \\ \hline
        HM LI       & \decoderHMLightenMHD  & \includestrip{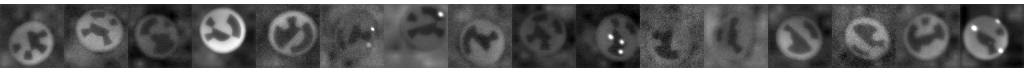}     \\ \hline
        HM BG       & \decoderHMBGMHD       & \includestrip{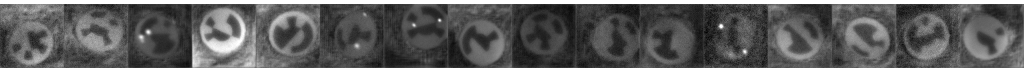}          \\ \hline
        RenderGAN   & \decoderRenderGANMHD  & \includestrip{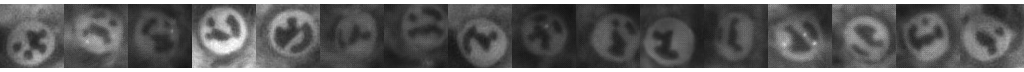}      \\ \hline
        RenderGAN + Real & \decoderRealRenderGANMHD  & \includestrip{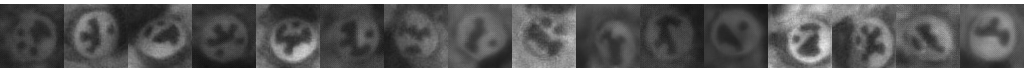} \\ \hline
        CV & \CVMHD
    \end{tabular}
\end{table}

\begin{figure}[t]
    \centering
    \input{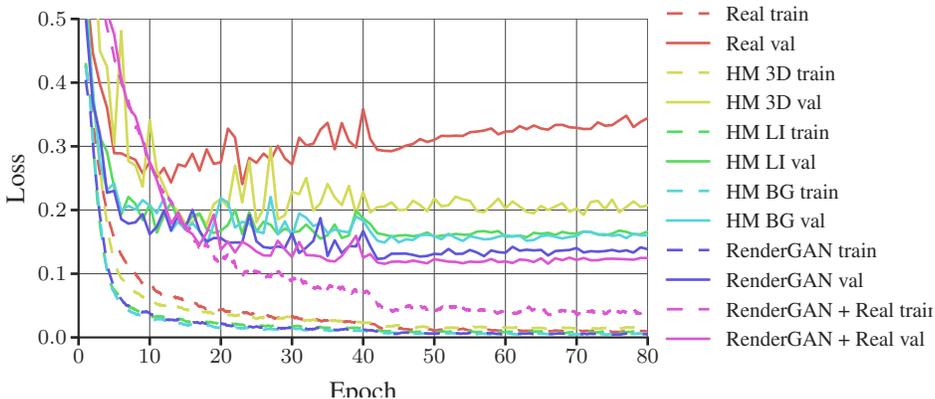}
    \caption{Training and validation losses of DCNNs trained on different data sets.
        As some data sets are missing the orientation of the tags, only the loss
        of the bits are plotted. Binary crossentropy is used as loss for the bits.
        The train and validation loss of each dataset have the same color.
    }
    \label{fig:decoder_hist}
\end{figure}

\textbf{Results:}
In Table~\ref{tab:mhd}, we present the results of the evaluation.
The training losses of the networks are plotted in Fig.~\ref{fig:decoder_hist}.
The model trained with the data generated by the RenderGAN has an MHD of \decoderRenderGANMHD.
The performance can furthermore be slightly improved by combining the generated with real data.
The small gap in performance when adding real data is a further indicator of the
quality of the generated samples.

If we use predictions from this DCNN instead of the computer vision pipeline,
the accuracy of the tracking improves from 55\% of the ids assigned correctly to 96\%.
At this quality, it is possible to analyze the social behavior of the honeybees reliably.

Compared to the handmade augmentations (\emph{HM 3D}), data from the RenderGAN
leads to considerably better performance.
The large gap in performance between the HM 3D and HM LI data highlights the
importance of the learned lighting augmentation.

\section{Discussion}
\label{sec:disscussion}

We proposed a novel extension to the GAN framework that is capable of rendering
samples from a basic 3D model more realistic.  Compared to computer graphics
pipelines, the RenderGAN can learn complex effects from unlabeled data that
would be otherwise hard to model with explicit rules.

Contrary to conventional GANs, the generator provides explicit information about
the synthesized images, which can be used as labels for a supervised algorithm.
The training of the RenderGAN requires no labels.

We showed an application of the RenderGAN framework to the BeesBook project, in
which the generator adds blur, lighting, background, and details to images
from a basic 3D model. The generated data looks strikingly real and includes
fine details such as spotlights, compression artifacts, and sensor noise.

In contrast to previous work that applied 3D models to produce training samples for DCNNs
\citep{Su2015, Richter2016ECCV, RosCVPR16}, we were able to train a DCNN from scratch with only
generated data that still generalizes to unseen real data.

While some work is required to adapt the RenderGAN to a specific domain,
once set up, arbitrary amounts of labeled data can be acquired cheaply, even if
the data distribution changes.  For example, if the tag design changes to
include more bits, small adaptions to the 3D model's source code and eventually the
hyperparameters of the augmentation functions would be sufficient. However, if
we had labeled the data by hand, then we would have to annotate data again.

While the proposed augmentations represent common image characteristics, a
disadvantage of the RenderGAN framework is that these augmentation functions
must be carefully customized for the application at hand to ensure that
high-level information is preserved. Furthermore, a suitable 3D model must be
available.

\section{Future Work}
\label{sec:futureWork}

For future work, it would be interesting to see the RenderGAN framework used on
other tasks where basic 3D models exist e.g.\ human
faces, pose estimation, or viewpoint prediction. In this context, one
could come up with different augmentation functions e.g.\ colorization, affine
transformations, or diffeomorphism. The RenderGAN could be especially valuable
to domains where pre-trained models are not available or when the annotations are
very complex. Another direction of future work might be to extend the RenderGAN
framework to other fields. For example, in speech synthesis, one could use an
existing software as a basic model and improve the realism of the output with
a similar approach as in the RenderGAN framework.

\newpage

\bibliographystyle{latex_sty/iclr2017_conference}
\bibliography{bibliography}

\begin{thebibliography}{25}
\providecommand{\natexlab}[1]{#1}
\providecommand{\url}[1]{\texttt{#1}}
\expandafter\ifx\csname urlstyle\endcsname\relax
  \providecommand{\doi}[1]{doi: #1}\else
  \providecommand{\doi}{doi: \begingroup \urlstyle{rm}\Url}\fi

\bibitem[Burt \& Adelson(1983)Burt and Adelson]{Burt1983}
Peter Burt and Edward Adelson.
\newblock The laplacian pyramid as a compact image code.
\newblock \emph{IEEE Transactions on communications}, 31\penalty0 (4):\penalty0
  532--540, 1983.

\bibitem[Chen et~al.(2016)Chen, Duan, Houthooft, Schulman, Sutskever, and
  Abbeel]{Chen2016}
Xi~Chen, Yan Duan, Rein Houthooft, John Schulman, Ilya Sutskever, and Pieter
  Abbeel.
\newblock Infogan: Interpretable representation learning by information
  maximizing generative adversarial nets.
\newblock \emph{arXiv preprint arXiv:1606.03657}, 2016.

\bibitem[Deng et~al.(2009)Deng, Dong, Socher, Li, Li, and Fei-Fei]{Deng2009}
Jia Deng, Wei Dong, Richard Socher, Li-Jia Li, Kai Li, and Li~Fei-Fei.
\newblock Imagenet: A large-scale hierarchical image database.
\newblock In \emph{Computer Vision and Pattern Recognition, 2009. CVPR 2009.
  IEEE Conference on}, pp.\  248--255. IEEE, 2009.

\bibitem[Denton et~al.(2015)Denton, Chintala, Fergus, et~al.]{Denton2015}
Emily~L Denton, Soumith Chintala, Rob Fergus, et~al.
\newblock Deep generative image models using a laplacian pyramid of adversarial
  networks.
\newblock In \emph{Advances in neural information processing systems}, pp.\
  1486--1494, 2015.

\bibitem[Dosovitskiy et~al.(2015)Dosovitskiy, Tobias~Springenberg, and
  Brox]{Dosovitskiy2014}
Alexey Dosovitskiy, Jost Tobias~Springenberg, and Thomas Brox.
\newblock Learning to generate chairs with convolutional neural networks.
\newblock In \emph{Proceedings of the IEEE Conference on Computer Vision and
  Pattern Recognition}, pp.\  1538--1546, 2015.

\bibitem[Girshick et~al.(2014)Girshick, Donahue, Darrell, and Malik]{Malik2014}
Ross Girshick, Jeff Donahue, Trevor Darrell, and Jitendra Malik.
\newblock Rich feature hierarchies for accurate object detection and semantic
  segmentation.
\newblock In \emph{Proceedings of the IEEE conference on computer vision and
  pattern recognition}, pp.\  580--587, 2014.

\bibitem[Goodfellow et~al.(2014)Goodfellow, Pouget-Abadie, Mirza, Xu,
  Warde-Farley, Ozair, Courville, and Bengio]{Goodfellow2014}
Ian Goodfellow, Jean Pouget-Abadie, Mehdi Mirza, Bing Xu, David Warde-Farley,
  Sherjil Ozair, Aaron Courville, and Yoshua Bengio.
\newblock Generative adversarial nets.
\newblock In \emph{Advances in Neural Information Processing Systems}, pp.\
  2672--2680, 2014.

\bibitem[Goodfellow et~al.(2016)Goodfellow, Bengio, and
  Courville]{Goodfellow2016}
Ian Goodfellow, Yoshua Bengio, and Aaron Courville.
\newblock Deep learning.
\newblock Book in preparation for MIT Press, 2016.
\newblock URL \url{http://www.deeplearningbook.org}.

\bibitem[He et~al.(2015)He, Zhang, Ren, and Sun]{He2015}
Kaiming He, Xiangyu Zhang, Shaoqing Ren, and Jian Sun.
\newblock Deep residual learning for image recognition.
\newblock \emph{arXiv preprint arXiv:1512.03385}, 2015.

\bibitem[Kingma \& Ba(2014)Kingma and Ba]{Kingma2014}
Diederik Kingma and Jimmy Ba.
\newblock Adam: A method for stochastic optimization.
\newblock \emph{arXiv preprint arXiv:1412.6980}, 2014.

\bibitem[Krizhevsky(2009)]{Krizhevsky2009}
Alex Krizhevsky.
\newblock {Learning Multiple Layers of Features from Tiny Images}, 2009.
\newblock ISSN 1098-6596.

\bibitem[Krizhevsky et~al.(2012)Krizhevsky, Sutskever, and
  Hinton]{Krizhevsky2012}
Alex Krizhevsky, Ilya Sutskever, and Geoffrey~E Hinton.
\newblock Imagenet classification with deep convolutional neural networks.
\newblock In \emph{Advances in neural information processing systems}, pp.\
  1097--1105, 2012.

\bibitem[Long et~al.(2015)Long, Shelhamer, and Darrell]{Long2015}
Jonathan Long, Evan Shelhamer, and Trevor Darrell.
\newblock {Fully Convolutional Networks for Semantic Segmentation}.
\newblock \emph{Proceedings of the IEEE Conference on Computer Vision and
  Pattern Recognition}, pp.\  3431--3440, 2015.
\newblock ISSN 10636919.
\newblock \doi{10.1109/CVPR.2015.7298965}.

\bibitem[Massa et~al.(2015)Massa, Russell, and Aubry]{Massa2015}
Francisco Massa, Bryan Russell, and Mathieu Aubry.
\newblock Deep exemplar 2d-3d detection by adapting from real to rendered
  views.
\newblock \emph{arXiv preprint arXiv:1512.02497}, 2015.

\bibitem[Peng et~al.(2015)Peng, Sun, Ali, and Saenko]{Peng2015}
Xingchao Peng, Baochen Sun, Karim Ali, and Kate Saenko.
\newblock {Learning Deep Object Detectors from 3D Models}.
\newblock \emph{Iccv}, 2015.
\newblock \doi{10.1109/ICCV.2015.151}.

\bibitem[Radford et~al.(2015)Radford, Metz, and Chintala]{Radford2015}
Alec Radford, Luke Metz, and Soumith Chintala.
\newblock {Unsupervised Representation Learning with Deep Convolutional
  Generative Adversarial Networks}, 2015.

\bibitem[Razavian et~al.(2014)Razavian, Azizpour, Sullivan, and
  Carlsson]{Razavian2014}
Ali~Sharif Razavian, Hossein Azizpour, Josephine Sullivan, and Stefan Carlsson.
\newblock Cnn features off-the-shelf: an astounding baseline for recognition.
\newblock In \emph{Proceedings of the IEEE Conference on Computer Vision and
  Pattern Recognition Workshops}, pp.\  806--813, 2014.

\bibitem[Richter et~al.(2016)Richter, Vineet, Roth, and
  Koltun]{Richter2016ECCV}
Stephan~R Richter, Vibhav Vineet, Stefan Roth, and Vladlen Koltun.
\newblock Playing for data: Ground truth from computer games.
\newblock In \emph{European Conference on Computer Vision}, pp.\  102--118.
  Springer, 2016.

\bibitem[Ros et~al.(2016)Ros, Sellart, Materzynska, Vazquez, and
  Lopez]{RosCVPR16}
German Ros, Laura Sellart, Joanna Materzynska, David Vazquez, and Antonio~M
  Lopez.
\newblock The synthia dataset: A large collection of synthetic images for
  semantic segmentation of urban scenes.
\newblock In \emph{Proceedings of the IEEE Conference on Computer Vision and
  Pattern Recognition}, pp.\  3234--3243, 2016.

\bibitem[Springenberg(2015)]{Springenberg2015}
Jost~Tobias Springenberg.
\newblock Unsupervised and semi-supervised learning with categorical generative
  adversarial networks.
\newblock \emph{arXiv preprint arXiv:1511.06390}, 2015.

\bibitem[Su et~al.(2015)Su, Qi, Li, and Guibas]{Su2015}
Hao Su, Charles~R Qi, Yangyan Li, and Leonidas~J Guibas.
\newblock Render for cnn: Viewpoint estimation in images using cnns trained
  with rendered 3d model views.
\newblock In \emph{Proceedings of the IEEE International Conference on Computer
  Vision}, pp.\  2686--2694, 2015.

\bibitem[Wang \& Gupta(2016)Wang and Gupta]{Wang2016}
Xiaolong Wang and Abhinav Gupta.
\newblock Generative image modeling using style and structure adversarial
  networks.
\newblock \emph{arXiv preprint arXiv:1603.05631}, 2016.

\bibitem[Wario et~al.(2015)Wario, Wild, Couvillon, Rojas, and
  Landgraf]{Wario2015}
Fernando Wario, Benjamin Wild, Margaret~Jane Couvillon, Ra{\'u}l Rojas, and Tim
  Landgraf.
\newblock Automatic methods for long-term tracking and the detection and
  decoding of communication dances in honeybees.
\newblock \emph{Frontiers in Ecology and Evolution}, 3:\penalty0 103, 2015.

\bibitem[Xiang et~al.(2014)Xiang, Mottaghi, and Savarese]{Xiang2014}
Yu~Xiang, Roozbeh Mottaghi, and Silvio Savarese.
\newblock Beyond pascal: A benchmark for 3d object detection in the wild.
\newblock In \emph{IEEE Winter Conference on Applications of Computer Vision},
  pp.\  75--82. IEEE, 2014.

\bibitem[Yosinski et~al.(2014)Yosinski, Clune, Bengio, and
  Lipson]{Yosinski2014}
Jason Yosinski, Jeff Clune, Yoshua Bengio, and Hod Lipson.
\newblock How transferable are features in deep neural networks?
\newblock In \emph{Advances in neural information processing systems}, pp.\
  3320--3328, 2014.

\end{thebibliography}

\newpage

\begin{appendices}

\section{Generated Images}
\label{appendix:generated_images}

\begin{figure}[h]
    \begin{subfigure}[b]{\textwidth}
        \centering
        \includegraphics[width=\linewidth]{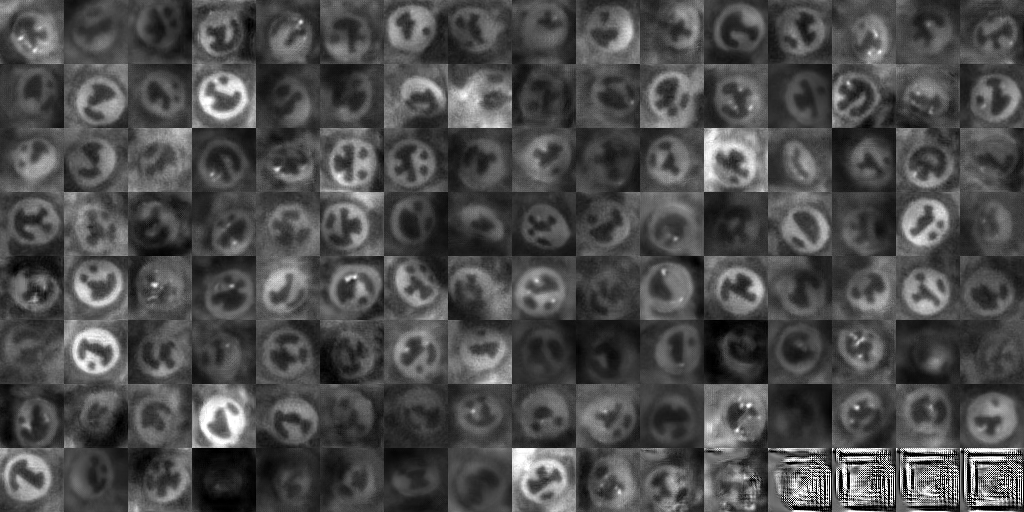}
        \caption{Generated images}
        \label{fig:tile_fake}
    \end{subfigure}
    \begin{subfigure}[b]{\textwidth}
        \centering
        \includegraphics[width=\linewidth]{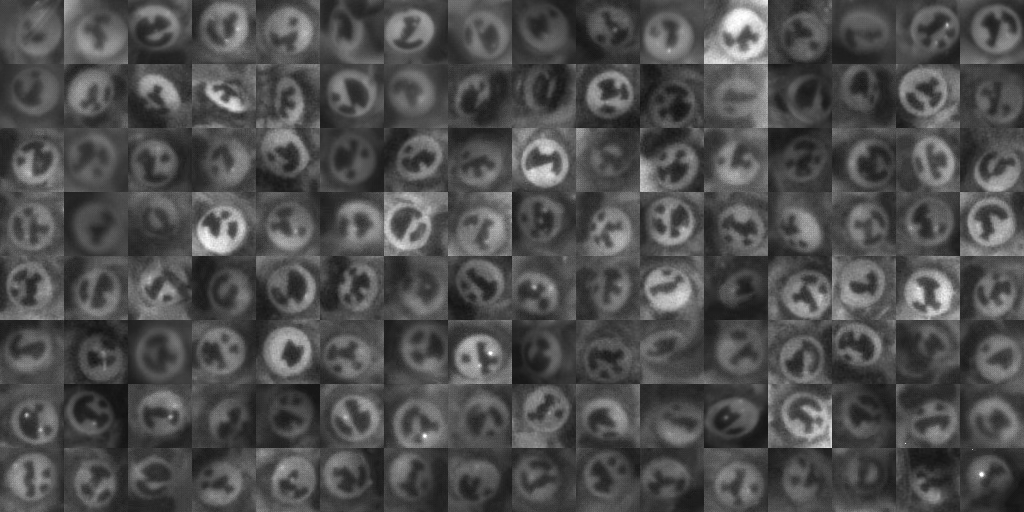}
        \caption{Real images}
        \label{fig:tile_real}
    \end{subfigure}
    \centering
    \caption{Continuum visualization on the basis of the discriminator score:
    Most realistc scored samples top left corner to least realistc bottom right
    corner. Images with artifacts are scored unrealistic and are not used for training.
    }
    \label{fig:}
\end{figure}

\newpage

\begin{figure}[h]
    \centering
    \includegraphics[width=\linewidth]{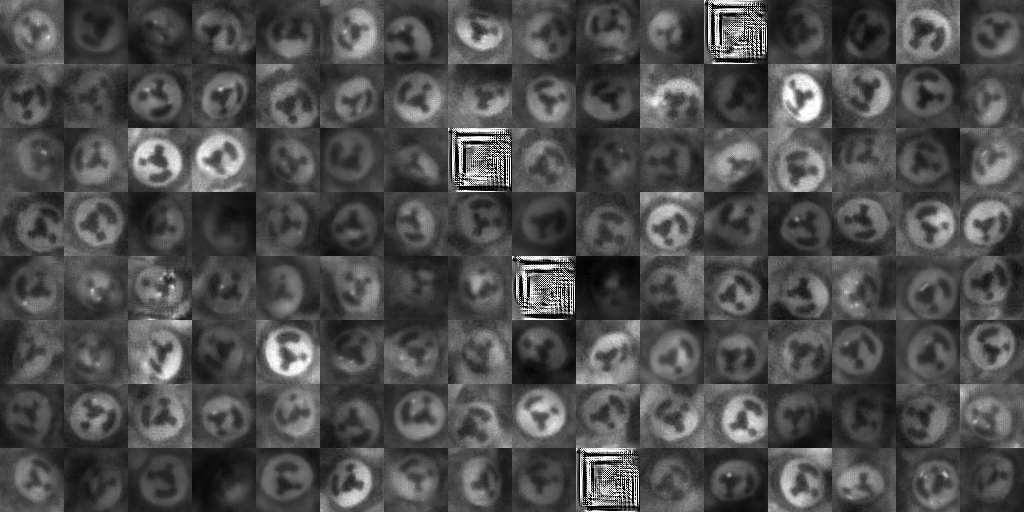}
    \caption{Images generated with the generator given a fixed bit configuration}
    \label{fig:tag3d_fake}
\end{figure}

\begin{figure}[h]
    \centering
    \includegraphics[width=\linewidth]{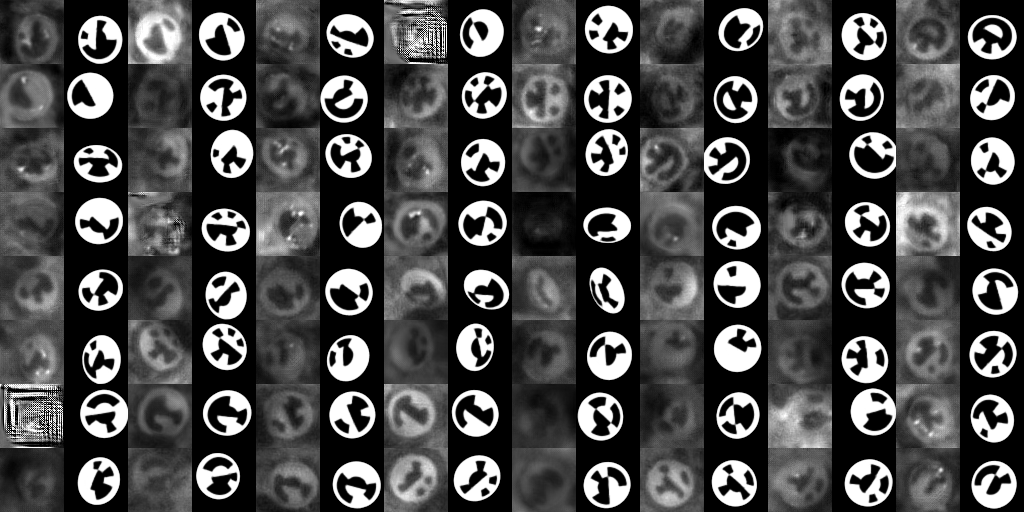}
    \caption{Correspondence of generated images and 3D model}
    \label{fig:tag3d_fake}
\end{figure}


\newpage

\section{Handmade augmentations}
\label{appendix:handmade}

We constructed augmentations for blur, lighting, background, noise and spotlights manually.
For synthesizing lighting, background and noise, we use image
    pyramids, i.e. a set of images $L_0, \ldots, L_6$ of size $(2^i
    \times 2^i)$ for $ 0 \le i \le 6$. Each level $L_i$ in the
pyramid is weighted by a scalar $\omega_i$.
Each pixel of the different level $L_i $  is drawn from $\mathcal{N}(0, 1)$.
The generated image $I_6$ is given by:
\begin{align}
    I_0 &= \omega_0 L_0 \\
    I_i &= \omega_i L_i + \text{upscale}(I_{i-1})
\end{align}
, where upscale doubles the image dimensions. The pyramid enables us to
generate random images while controlling their frequency domain by
weighting the pyramid levels appropriately.

\begin{itemize}
    \item \textbf{Blur:} Gaussian blur with randomly sampled scale.
    \item \textbf{Lighting: } Similar as in the RenderGAN. Here, the scaling of the white
        and black parts and shifting is constructed with image pyramids.
    \item \textbf{Background: } image pyramids with the lower levels weight more.
    \item \textbf{Noise:} image pyramids with only the last two layer.
    \item \textbf{Spotlights}: overlay with possible multiple 2D
        Gauss function with a random position on the tag and random covariance.
\end{itemize}

We selected all parameters manually by comparing the generated to real images.
However, using slightly more unrealistic images resulted in better performance
of the DCNN trained with the HM 3D data. The parameters of the handmade augmentations
can be found online in our source code repository.

\section{Augmentations of the real data}
\label{appendix:augmentation_real}

We scale and shift the pixel intensities randomly, i.e. $ s I + t $,
where $I$ is the image and $s$, $t$ are scalars.
The noise is sampled for each pixel from $\mathcal{N}(0, \epsilon )$,
where $\epsilon \sim \max(0,~\mathcal{N}(\mu_n, \sigma_n))$ is drawn for each image separately.
Different affine transformations (rotation, scale, translation, and shear) are used.
\begin{table}[h]
\caption{Parameters of the augmentation of the real data}
\label{tab:augmentation_real}
\begin{center}
    \begin{tabular}{c | c}
        Name & Distribution \\ \hline
        Intensity Scale (s) & unif(0.9, 1.1) \\
        Intensity Shift (t) & unif(-0.2, 0.2) \\
        Noise Mean ($\mu_n$) & 0.04 \\
        Noise Std ($\sigma_n$)& 0.03 \\
        Rotation & unif(0, 2$\pi$) \\
        Scale & unif(0.7, 1.1) \\
        Shear & unif(-0.3, 0.3) \\
        Translation & unif(-4, 4) \\
    \end{tabular}
\end{center}
\end{table}

\newpage

\section{Training samples}
\label{appendix:training_samples}
\begin{figure}[h]
    \label{fig:datasets_appendix}
    \centering
    \begin{subfigure}[h]{\textwidth}
        \includegraphics[width=\linewidth]{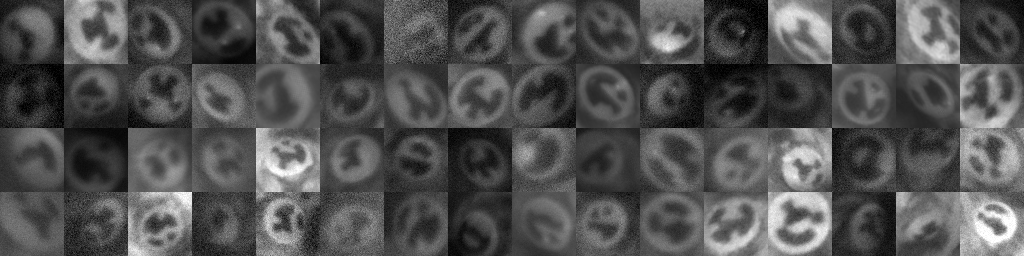}
        \caption{Real trainings samples}
    \end{subfigure}
    \begin{subfigure}[h]{\textwidth}
        \includegraphics[width=\linewidth]{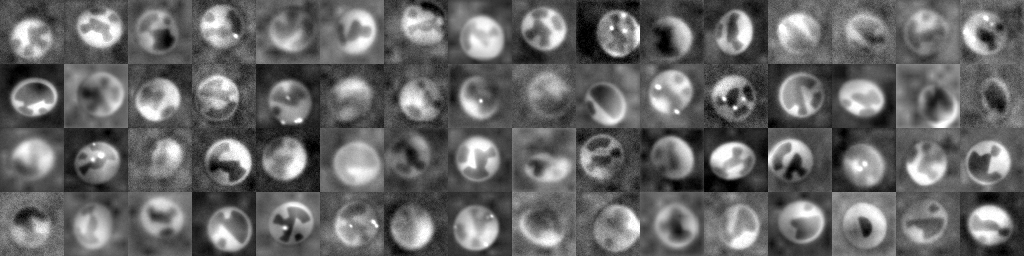}
        \caption{HM 3D training samples}
    \end{subfigure}
    \begin{subfigure}[h]{\textwidth}
        \includegraphics[width=\linewidth]{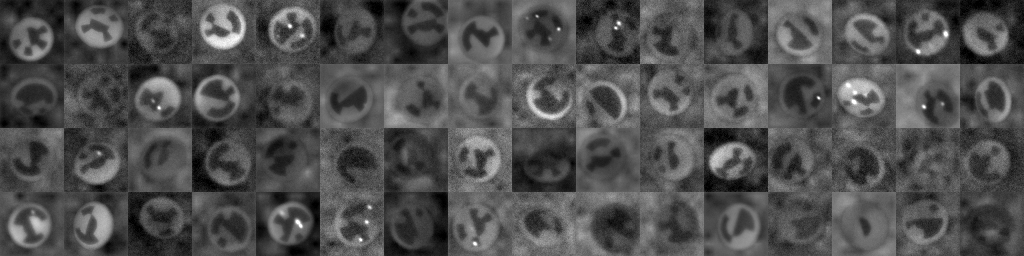}
        \caption{HM LI training samples}
    \end{subfigure}
    \begin{subfigure}[h]{\textwidth}
        \includegraphics[width=\linewidth]{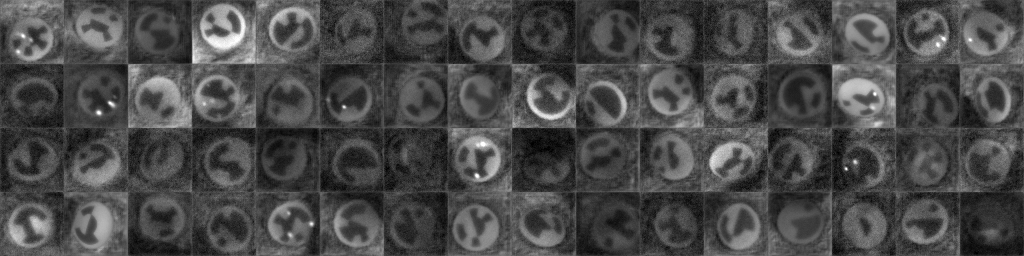}
        \caption{HM BG training samples}
    \end{subfigure}
    \begin{subfigure}[h]{\textwidth}
        \includegraphics[width=\linewidth]{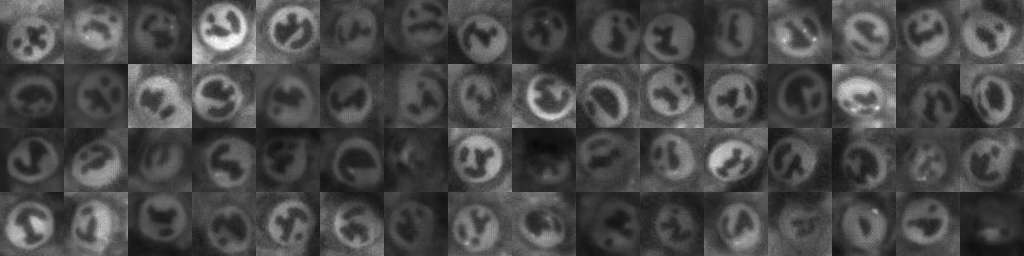}
        \caption{RenderGAN}
    \end{subfigure}
    \caption{Training samples from the different datasets.}
\end{figure}

\end{appendices}

\end{document}